\documentclass[11pt]{article}
\usepackage{eacl2017}
\usepackage{times}
\usepackage{url}
\usepackage{latexsym}
\usepackage{xspace}
\usepackage{multirow}
\usepackage[nomessages]{fp}

\def\roundposition{1}

\edef\rounded{0}
\newcommand{\rdm}[1]{\edef\rounded{0}\FPeval\rounded{round(#1,\roundposition)}\rounded}

\newcommand\BLEU{\textsc{Bleu}\xspace}
\newcommand\TER{\textsc{Ter}\xspace}

\eaclfinalcopy 

\title{Ensemble Distillation for Neural Machine Translation}

\author{Markus Freitag, Yaser Al-Onaizan and Baskaran Sankaran\\
           IBM T.J. Watson Research Center\\
            1101 Kitchawan Rd, Yorktown Heights, NY 10598\\
        {\{freitagm,onaizan,bsankaran\}}@us.ibm.com}

\date{}

\begin{document}
\maketitle
\begin{abstract}
Knowledge distillation describes a method for training a student network to 
perform better by learning from a stronger teacher network. Translating a sentence
with an Neural Machine Translation (NMT) engine is time expensive and having a smaller
model speeds up this process.
We demonstrate how to transfer the translation quality of an ensemble and an oracle \BLEU teacher network into a single NMT system. Further, we present translation improvements 
from a teacher network that has the same architecture and dimensions of the student network.
As the training of the student model is still expensive, we introduce a data filtering method
based on the knowledge of the teacher model that not only speeds up the training, but also
leads to better translation quality.
Our techniques need no code change and can be easily reproduced with any NMT architecture to
speed up the decoding process.

\end{abstract}

\section{Introduction}
Knowledge distillation describes the idea of enhancing a student network by matching its predictions to the ones of a stronger teacher network.
There are two possible ways of using knowledge distillation for Neural Machine Translation (NMT): First, the student network can be
a model with less layers and/or hidden units. The main purpose of this is to reduce the model size of the NMT system
without significant loss in translation quality. Secondly, without changing the model architecture, one can get reasonable gains
by combining different models of the same architecture with ensemble. By using an ensemble, you also get the disadvantage of a 
much slower decoding speed. 

We show that the performance of a teacher composed of an ensemble of 6 models can be achieved with a student composed of a single model leading to a significantly faster decoding and smaller memory footprint.
We also investigate a teacher network that is producing the oracle \BLEU translations from the final decoder beam and demonstrate how to improve an NMT system, even if the student network has the same architecture and dimensions as the teacher network. 
In our knowledge distillation approach, we translate the full training data with the teacher model to use the translations as additional training data for the student network. This kind of knowledge transfer does not need any source code modification and can be reproduced by any NMT network architecture.

Training an NMT system on several millions of parallel sentences is already slow. When applying knowledge distillation,
a second training phase on at least the same amount of data is needed. We show how to use the knowledge of the teacher model to filter the training data.
We show that filtering the data does not only make the training faster, but also improves the translation quality.

To summarize the main contributions:

\begin{itemize}

    \item We apply knowledge distillation on an ensemble and oracle \BLEU teacher model
    \item We demonstrate how to successfully use knowledge distillation if the student network has the same architecture and dimensions as the teacher network
    \item We introduce a simple and easy reproducible approach
    \item We filter the training data with the knowledge of the teacher model
    \item We compare different parameter initializations for the student network
\end{itemize}

\section{Knowledge Distillation}

The idea of knowledge distillation is to match the predictions of a student network to that of a teacher network.
In this work, we collect the predictions of the teacher network by translating the full training data with the teacher network. 
By doing this, we produce a new reference for the training data which can be used by the student network to simulate the
teacher network.  
There are two ways of using the forward translation. First, we can train the student network only on the original source and the translations.
Secondly, we can add the translations as additional training data to the original training data. This has the side effect that
the final training data size of the student network is doubled. 

\section{Teacher Networks}

\begin{itemize}

    \item \textbf{Ensemble Teacher Model} \\
        An ensemble of different NMT models can improve the translation performance of an NMT system.
        The idea is to train several models in parallel and combine their predictions
        by averaging the probabilities of the individual models at each time step during decoding.
        In this work, we use an ensemble of 6 models as a teacher model. All 6 individual systems are trained
        on the same parallel data and use the same optimization method. 
        The only difference is the random initialization of the parameters.

    \item \textbf{Oracle \BLEU Teacher Model} \\
       We use a left-to-right beam-search decoder to build new translations that aims to maximize the conditional probability of a given model.
       It stops the search when it found a fix number of hypotheses that end with an end-of-sequence symbol 
       and picks the translation with the highest
       log-probability out of the final candidate list. 
       In our distillation approach, we produce the forward translation of our parallel data. Since we know the reference
       translation of all sentences, we choose instead of the highest log-probability the sentence with the highest
       sentence level \BLEU from the final candidate list. We use the sentence level \BLEU proposed by \cite{lin2004automatic}
       which adds 1 to both the matched and the total n-gram counts.
\end{itemize}

\section{Data Filtering}

In machine translation, bilingual sentence pairs that serve as training data are mostly crawled from the web and contain many nonparallel sentence pairs.
Furthermore, one source sentence can have several correct translations that differ in choice and order of words.
The training of the network gets complicated, if the training corpus contains noisy sentence pairs or sentences with several correct translations.
In our knowledge distillation approach, we translate the full parallel data with our teacher model. This gives us the option to 
score each translation with the original reference. We remove sentences with high \TER scores \cite{snover2006study} from our training
data. By removing noisy or unreachable sentence pairs, the training algorithm is able to learn a stronger network.

\section{Experiments}
\label{sec:experiments}
We run our experiments on the German$\rightarrow$English WMT 2016 translation task \cite{bojar2016findings} (3.9M parallel sentences) and use newstest2014 as validation and newstest2015 as test set. 
We use our in-house attention-based NMT implementation which is similar to ~\cite{bahdanau+:2014}.
We use sub-word units extracted by byte pair encoding~\cite{sennrich2015neural} instead of words
which shrinks the vocabulary to 40k sub-word symbols for both source and target. 
We use an embedding dimension of 620 and fix the RNN GRU layers to be of 1000 cells each. 
For the training procedure, we use SGD to update the model parameters with a mini-batch size of 64.
Starting from the 4th epoch, we reduce the learning rate by half every epoch.
The training data is shuffled after each epoch and we use beam 5 for all translations.
We stopped the training of each setup (including baseline) when the validation score did not improve
in the last 3 epochs.
All different setups are run twice: First, we train the student network from scratch with random parameter initialization.
Secondly, we continue the training based on the final parameters of the baseline model.
\section{Results}
\begin{itemize}

    \item \textbf{Single Teacher Model} \\
       Instead of using a stronger teacher model, we use the same model for both student and teacher network.
       By using the forward translation, we can stabilize the student network and make its decision much stronger.
       Results are given in Table~\ref{tab:deen-results-single}. Using only the forward translation does not improve
       the model. When combining both the reference and the forward translation, we improve the model by 1.4 points in both \BLEU
       and \TER. Pruning the training data and using only sentence pairs with a \TER score less than 0.8
       yields to similar translation quality while reducing the training data by 12\% leading to a faster training.

    \item \textbf{Ensemble Teacher Model} \\
       The results for using an ensemble of 6 models as a teacher model are summarized in Table~\ref{tab:deen-results-ensemble}.
       Using only the forward translation improves the single system by 1.4 points in \BLEU and 1.9 points in \TER.
       When using both the original reference and the forward translation, we get an additional improvement of 0.3 points in \BLEU.
       When pruning the parallel data and using only sentences with a \TER less than 0.8, we can improve the single system by 2 points in \BLEU and 2.2 points in \TER.

    \item \textbf{Oracle \BLEU Teacher Model} \\
        The teacher model is the same ensemble model as before, but instead of choosing the hypotheses with the highest log probability, it chooses the sentence with the highest sentence level \BLEU from the final candidate list.
        The empirical results obtained with the oracle \BLEU teacher model are summarized in Table~\ref{tab:deen-results-sbleu}. 
        By using only the forward translation of the teacher network, we gain improvements of 1.1 points in \BLEU and 1.2 points in \TER.
        By combining both the forward translation and the reference, we obtain improvements of 1.5 points in both \BLEU and \TER.
        However, the results are slightly worse compared to the results obtained with an ensemble teacher network.

    \item \textbf{Reducing Model Size} \\
        We use the ensemble teacher network to teach a student network with lower dimensions. Empirical results are given in
        Table~\ref{tab:deen-results-reduce}. We reduced the original word embedding (Wemb) size of 620 to 150 and the original hidden layer size (hlayer) of
        1000 to 300 without losing any translation quality compared to the single model. In fact the performance is even better by 0.4 points in \BLEU and
        0.6 points in \TER.

\end{itemize}

\begin{table*}[ht!]
    \centering
    \begin{tabular}{|l|c|c|c|c|c|c|c|c|}
        \hline
        setup & parallel data & from & continue & \multicolumn{2}{c}{\bf{newstest2014}} & \multicolumn{2}{c|}{\bf{newstest2015}}  \\
         & &scratch & training & \BLEU & \TER & \BLEU & \TER \\ \hline \hline
        baseline single & original (4M) & $\surd$ &  & \rdm{27.32} & \rdm{54.64} & \rdm{27.43} & \rdm{53.72} \\ \cline{1-4} \hline
        \multirow{4}{*}{distillation all} & \multirow{2}{2.5cm}{trans baseline (4M)} & $\surd$ & & \rdm{27.48} & \rdm{54.19} & \rdm{27.65} & \rdm{53.31} \\ \cline{3-8}
         &  & & $\surd$ & \rdm{27.36} & \rdm{54.63} & \rdm{27.48} & \rdm{53.37} \\ \cline{2-8}
         & \multirow{2}{2.5cm}{trans baseline + original (8M)} & $\surd$ & & \rdm{28.63} & \rdm{53.28} & \bf{\rdm{28.77}} & \bf{\rdm{52.28}} \\ \cline{3-8}
                                      & & & $\surd$ & \rdm{28.26} & \rdm{53.70} & \rdm{28.30} & \rdm{52.87} \\ \hline
        \multirow{6}{*}{distillation \TER $\leq$ 0.8} & \multirow{2}{*}{reference (3.5M)} & $\surd$ & & \rdm{26.95} & \rdm{55.00} & \rdm{27.36} & \rdm{53.76} \\ \cline{3-8}
         & & & $\surd$ & \rdm{27.56} & \rdm{54.59} & \rdm{27.70} & \rdm{53.62} \\ \cline{2-8}
         & \multirow{2}{2.5cm}{trans baseline (3.5M)} & $\surd$ & & \rdm{27.50} & \rdm{54.07} & \rdm{27.67} & \rdm{53.21} \\ \cline{3-8}
         & & & $\surd$ & \rdm{27.41} & \rdm{54.31} & \rdm{27.46} & \rdm{53.33} \\ \cline{2-8}
         & \multirow{2}{2.5cm}{trans baseline + original (7M)} & $\surd$ & & \rdm{28.61} & \rdm{52.95} & \rdm{28.49} & \rdm{52.38} \\ \cline{3-8}
         & & & $\surd$ & \rdm{28.50} & \rdm{53.32} & \bf{\rdm{28.58}} & \bf{\rdm{52.53}} \\ \hline
    \end{tabular}
    \caption{German-English: Knowledge distillation based on a single teacher model with same architecture and dimensions as the student networks.}
    \label{tab:deen-results-single}
\end{table*}

\begin{table*}[t!]
    \centering
    \begin{tabular}{|l|c|c|c|c|c|c|c|c|}
        \hline
        setup & parallel data & from & continue & \multicolumn{2}{c}{\bf{newstest2014}} & \multicolumn{2}{c|}{\bf{newstest2015}}  \\
         & &scratch & training & \BLEU & \TER & \BLEU & \TER \\ \hline \hline
        ensemble of 6 (teacher) & original (4M) & $\surd$ &  & \rdm{29.82} & \rdm{51.74} & \rdm{29.77} & \rdm{51.15} \\ \hline
        \multirow{4}{*}{distillation} & \multirow{2}{*}{trans ens (4M)} & $\surd$ & & \rdm{28.58} & \rdm{52.90} & \rdm{28.43} & \rdm{52.41}\\ \cline{3-8}
                                      & & & $\surd$ & \rdm{28.80} & \rdm{52.72} & \rdm{28.80} & \rdm{51.76} \\ \cline{2-4} \cline{2-8}
                                      & \multirow{2}{2.5cm}{trans ens + original (8M)} & $\surd$ & & \rdm{29.10} & \rdm{52.58} & \rdm{29.03} & \rdm{51.96} \\ \cline{3-8}
                                      & & & $\surd$ & \rdm{28.92} & \rdm{52.61} & \bf{\rdm{29.09}} & \bf{\rdm{51.81}} \\ \hline
        \multirow{2}{*}{distillation \TER $\leq$ 0.8} & \multirow{2}{2.5cm}{trans ens + original (7M)} & $\surd$ & & \rdm{29.19} & \rdm{52.34} & \rdm{29.16} & \rdm{51.67} \\ \cline{3-8}
                                                      & & & $\surd$ & \rdm{29.29} & \rdm{52.40} & \bf{\rdm{29.35}} & \bf{\rdm{51.49}} \\ \hline
    \end{tabular}
    \caption{German-English: Knowledge distillation based on a ensemble teacher model.}
    \label{tab:deen-results-ensemble}
\end{table*}

\begin{table*}[t!]
    \centering
    \begin{tabular}{|l|c|c|c|c|c|c|c|c|}
        \hline
        setup & parallel data & from & continue & \multicolumn{2}{c}{\bf{newstest2014}} & \multicolumn{2}{c|}{\bf{newstest2015}}  \\
         & &scratch & training & \BLEU & \TER & \BLEU & \TER \\ \hline \hline
        o\BLEU (teacher) & original (4M) & $\surd$ &  & \rdm{34.49} & \rdm{46.90} & \rdm{33.74} & \rdm{46.38} \\ \hline
        \multirow{3}{*}{distillation} & o\BLEU trans (4M) & & $\surd$ & \rdm{28.52} & \rdm{53.14} & \rdm{28.51} & \rdm{52.46} \\ \cline{2-4} \cline{3-8}
                                      & \multirow{2}{3cm}{oracle \BLEU trans + original (8M)} & $\surd$ & & \rdm{28.86} & \rdm{52.75} & \bf{\rdm{28.87}} & \bf{\rdm{52.24}} \\ \cline{3-8}
                                      & & & $\surd$ & \rdm{28.90} & \rdm{52.98} & \rdm{28.71} & \rdm{52.30} \\ \hline
    \end{tabular}
    \caption{German-English: Knowledge distillation based on oracle \BLEU (o\BLEU) translations.}
    \label{tab:deen-results-sbleu}
\end{table*}

\begin{table*}[t!]
    \centering
    \begin{tabular}{|l|c|c|c|c|c|c|c|c|}
        \hline
        setup & parallel data & from & model size & \multicolumn{2}{c}{\bf{newstest2014}} & \multicolumn{2}{c|}{\bf{newstest2015}}  \\
              & & scratch & (hlayer,Wemb) & \BLEU & \TER & \BLEU & \TER \\ \hline \hline
        baseline single & original (4M) & & 1000,620 & \rdm{27.32} & \rdm{54.64} & \rdm{27.43} & \rdm{53.72} \\ \cline{1-4} \hline
        ensemble of 6 (teacher) & original (4M) & & 1000,620 & \rdm{29.82} & \rdm{51.74} & \rdm{29.77} & \rdm{51.15} \\ \hline \hline
        \multirow{4}{*}{distillation} & \multirow{4}{2.2cm}{trans ens + original (8M)} & $\surd$ & 1000,620 & \rdm{29.10} & \rdm{52.58} & \rdm{29.03} & \rdm{51.96} \\ \cline{3-8}
         & & $\surd$ & 750,400 & \rdm{28.98} & \rdm{52.65} & \rdm{28.90} & \rdm{52.04} \\ \cline{3-8}
         & & $\surd$ & 500,250 & \rdm{28.47} & \rdm{53.09} & \rdm{28.53} & \rdm{52.50} \\ \cline{3-8}
         & & $\surd$ & 300,150 & \rdm{27.56} & \rdm{54.04} & \rdm{27.77} & \rdm{53.07} \\ \hline
    \end{tabular}
    \caption{German-English model size reduction: Knowledge distillation with the ensemble teacher model. The student networks differ in the number of parameters used for both hidden layer and word embedding.}
    \vspace{0.4cm}
    \label{tab:deen-results-reduce}
\end{table*}


\section{Related Work}
\cite{Bucilua:2006:MC:1150402.1150464} show how to compress the function that
is learned by a complex ensemble model into a much smaller, faster
model that has comparable performance.
Results on eight test problems show that, on average, the loss in performance due
to compression is usually negligible.

\cite{ba2014deep} demonstrate that shallow feed-forward nets can learn
the complex functions previously learned by deep nets with knowledge distillation.
On the TIMIT phoneme recognition and CIFAR-10 image recognition tasks, shallow nets can be trained
that perform similarly to deeper convolutional models.

\cite{hinton2015distilling} present knowledge distillation for image classification (MNIST) and
acoustic modelling. They show that nearly all of the improvement that is achieved by training an
ensemble of deep neural nets can be distilled into a single neural net of the same size.

\cite{kim2016sequence} use knowledge distillation for NMT to reduce the model size of their neural network.
Their best student model runs 10 times faster with little loss in performance. Even their work is quite similar and in fact was the motivation for our work, there are several differences:

\begin {itemize}
    \item We run experiments with an ensemble teacher model whereas Kim and Rush only reduced the dimension of a single teacher network.
    \item Kim and Rush run experiments based on a combination of oracle \BLEU and forward translation. We instead successfully showed how to use only the oracle translation for the teacher model.
    \item We utilized both the forward translation and the original reference in our experiments which lead to reasonable improvements in comparison to only using the forward translation.
    \item In addition, we used the information from the forward translation and pruned the
training data which does not only speed up the training, but also improves the performance.
    \item We showed how to successfully use knowledge distillation to even benefit from a teacher network that has the same architecture and dimensions as the student network.
    \item We further investigate, if the parameters of the student model should be randomly initialize or 
if we start training from the final parameters of a baseline student network that has been trained on the
given parallel data only.
\end{itemize}

\section{Conclusion}

In this work, we applied knowledge distillation for several kinds of teacher networks. First we demonstrate how to benefit from
a teacher network that is the same architecture as the student network. By combining both forward translation and the original reference, we
get an improvement of 1.4 points in \BLEU. Using an ensemble model of 6 single models as teacher model further improves the
translation quality of the student network. We showed how to prune the parallel data based on the \TER values
obtained with the forward translations. The combination of an ensemble teacher network and pruning all sentences with a \TER value
higher than 0.8 leads us to the best setup which improves the baseline by 2 points in \BLEU and 2.2 points in \TER.
Using a teacher model based on the oracle \BLEU translations does improve the
translation quality, but the results are slightly worse compared to the ensemble teacher model.
Furthermore, we showed how to use the teacher ensemble model to significantly reduce the
size of the student network while still getting gains in translation quality. 


\bibliography{distillation}

\begin{thebibliography}{}

\bibitem[\protect\citename{Ba and Caruana}2014]{ba2014deep}
Jimmy Ba and Rich Caruana.
\newblock 2014.
\newblock Do deep nets really need to be deep?
\newblock In {\em Advances in neural information processing systems}, pages
  2654--2662.

\bibitem[\protect\citename{{Bahdanau} \bgroup et al.\egroup
  }2014]{bahdanau+:2014}
D.~{Bahdanau}, K.~{Cho}, and Y.~{Bengio}.
\newblock 2014.
\newblock {Neural Machine Translation by Jointly Learning to Align and
  Translate}.
\newblock {\em ArXiv e-prints}, September.

\bibitem[\protect\citename{Bojar \bgroup et al.\egroup
  }2016]{bojar2016findings}
Ondrej Bojar, Rajen Chatterjee, Christian Federmann, Yvette Graham, Barry
  Haddow, Matthias Huck, Antonio~Jimeno Yepes, Philipp Koehn, Varvara
  Logacheva, Christof Monz, et~al.
\newblock 2016.
\newblock Findings of the 2016 conference on machine translation (wmt16).
\newblock {\em Proceedings of WMT}.

\bibitem[\protect\citename{Bucilu\v{a} \bgroup et al.\egroup
  }2006]{Bucilua:2006:MC:1150402.1150464}
Cristian Bucilu\v{a}, Rich Caruana, and Alexandru Niculescu-Mizil.
\newblock 2006.
\newblock Model compression.
\newblock In {\em Proceedings of the 12th ACM SIGKDD International Conference
  on Knowledge Discovery and Data Mining}, KDD '06, pages 535--541, New York,
  NY, USA. ACM.

\bibitem[\protect\citename{Hinton \bgroup et al.\egroup
  }2015]{hinton2015distilling}
Geoffrey Hinton, Oriol Vinyals, and Jeff Dean.
\newblock 2015.
\newblock Distilling the knowledge in a neural network.
\newblock {\em arXiv preprint arXiv:1503.02531}.

\bibitem[\protect\citename{Kim and Rush}2016]{kim2016sequence}
Yoon Kim and Alexander~M Rush.
\newblock 2016.
\newblock Sequence-level knowledge distillation.
\newblock {\em arXiv preprint arXiv:1606.07947}.

\bibitem[\protect\citename{Lin and Och}2004]{lin2004automatic}
Chin-Yew Lin and Franz~Josef Och.
\newblock 2004.
\newblock Automatic evaluation of machine translation quality using longest
  common subsequence and skip-bigram statistics.
\newblock In {\em Proceedings of the 42nd Annual Meeting on Association for
  Computational Linguistics}, page 605. Association for Computational
  Linguistics.

\bibitem[\protect\citename{Sennrich \bgroup et al.\egroup
  }2015]{sennrich2015neural}
Rico Sennrich, Barry Haddow, and Alexandra Birch.
\newblock 2015.
\newblock Neural machine translation of rare words with subword units.
\newblock {\em arXiv preprint arXiv:1508.07909}.

\bibitem[\protect\citename{Snover \bgroup et al.\egroup }2006]{snover2006study}
Matthew Snover, Bonnie Dorr, Richard Schwartz, Linnea Micciulla, and John
  Makhoul.
\newblock 2006.
\newblock A study of translation edit rate with targeted human annotation.
\newblock In {\em Proceedings of association for machine translation in the
  Americas}, volume 200.

\end{thebibliography}
\bibliographystyle{distillation}

\end{document}